\pdfoutput=1

\documentclass[11pt]{article}

\usepackage[]{emnlp2021}

\usepackage{times}
\usepackage{latexsym}
\usepackage[T1]{fontenc}
\usepackage[utf8]{inputenc}

\usepackage{microtype}

\usepackage{url}
\usepackage{multirow}
\usepackage{amsmath}
\usepackage{amssymb}
\usepackage{graphicx}
\usepackage{colortbl}
\usepackage{subcaption}

\newcommand{\err}[1]{\small{$\pm$ #1}}

\newcommand{\graycell}{\cellcolor{lightgray!30}}
\newcommand{\graybox}[1]{\colorbox{lightgray!30}{#1}}
\newcommand{\resq}[1]{\textbf{RQ#1}}
\newcommand{\del}[1]{\small{(#1)}}
\definecolor{neutgreen}{RGB}{34,149,34}
\definecolor{neutred}{RGB}{255,0,0}
\newcommand{\incd}{(\textcolor{neutgreen}{$\uparrow$})}
\newcommand{\incdd}{(\textcolor{neutred}{$\uparrow$})}
\newcommand{\decd}{(\textcolor{neutred}{$\downarrow$})}
\newcommand{\decdd}{(\textcolor{neutgreen}{$\downarrow$})}
\newcommand{\neutd}{(\textcolor{gray}{--})}

\newcommand{\be}{\textsc{be}}
\newcommand{\bg}{\textsc{bg}}
\newcommand{\ga}{\textsc{ga}}
\newcommand{\mhr}{\textsc{mhr}}
\newcommand{\mt}{\textsc{mt}}
\newcommand{\ug}{\textsc{ug}}
\newcommand{\ur}{\textsc{ur}}
\newcommand{\vi}{\textsc{vi}}
\newcommand{\wo}{\textsc{wo}}

\newcommand{\fttrained}{\textsc{fastT}}
\newcommand{\berttrained}{\textsc{bert}}
\newcommand{\mbert}{\textsc{mBert}}

\newcommand{\lapt}{\textsc{lapt}}
\newcommand{\va}{\textsc{va}}
\newcommand{\tva}{\textsc{tva}}

\title{Specializing Multilingual Language Models: An Empirical Study}
\author{Ethan C. Chau$^{\dagger}$ \quad
  Noah A. Smith$^{\dagger \star}$ \\
  $^\dagger$Paul G. Allen School of Computer Science \& Engineering, University of Washington \\
  $^\star$Allen Institute for Artificial Intelligence \\
  {\tt \{echau18,nasmith\}@cs.washington.edu}}

\date{}

\begin{document}
\maketitle
\begin{abstract}
    Pretrained multilingual language models have become a common tool in transferring NLP capabilities to low-resource languages, often with adaptations.  In this work, we study the performance, extensibility, and interaction of two such adaptations: vocabulary augmentation and script transliteration.  Our evaluations on part-of-speech tagging, universal dependency parsing, and named entity recognition in nine diverse low-resource languages uphold the viability of these approaches while raising new questions around how to optimally adapt multilingual models to low-resource settings.
\end{abstract}

\section{Introduction}

Research in natural language processing is increasingly carried out in languages beyond English.  This includes \emph{high-resource} languages with abundant data, as well as \emph{low-resource} languages, for which labeled (and unlabeled) data is scarce.  In fact, many of the world's languages fall into the latter category, even some with a high number of speakers.  This presents unique challenges compared to high-resource languages: effectively modeling low-resource languages involves both accurately tokenizing text in such languages and maximally leveraging the limited available data.

One common approach to low-resource NLP is the \emph{multilingual} paradigm, in which methods that have shown success in English are applied to the union of many languages' data,\footnote{Within the multilingual paradigm, a distinction is sometimes made between \emph{massively multilingual} methods, which consider tens or hundreds of languages; and \emph{polyglot} methods, which use only a handful.  In this paper, all mentions of ``multilingual'' refer to the former.} enabling transfer between languages.  For instance, multilingual contextual word representations (CWRs) from language models \citep[][\emph{inter alia}]{devlin-etal-2019-bert,huang-etal-2019-unicoder,lample2019cross} are conventionally ``pretrained'' on large multilingual corpora before being ``finetuned'' directly on supervised tasks; this pretraining-finetuning approach is derived from analogous monolingual models \citep{devlin-etal-2019-bert,liu2019roberta,peters-etal-2018-deep}.  However, considering the diversity of the world's languages and the great data imbalance among them, it is natural to question whether the current multilingual paradigm can be improved upon for low-resource languages.

Indeed, past work has demonstrated that it can.  For instance, \citet{wu-dredze-2020-languages} find that multilingual models often lag behind non-contextualized baselines for the lowest-resource languages in their training data, drawing into question their utility in such settings.  \citet{conneau-etal-2020-unsupervised} posit that this phenomenon is a result of limited model capacity, which proves to be a bottleneck for sufficient transfer to low-resource languages.  In fact, with multilingual models only being pretrained on a limited set of languages, most of the world's languages are unseen by the model.  For such languages, the performance of such models is even worse \citep{chau-etal-2020-parsing}, due in part to the diversity of scripts across the world's languages \citep{muller2020unseen,pfeiffer2021unks,rust-etal-2021-good} as compared to the models' Latin-centricity \citep{acs-2019-exploring}.

Nonetheless, there have been multiple attempts to remedy this discrepancy by \emph{specializing}\footnote{We use \emph{specialization} to denote preparing a model for use on a specific target language, to the exclusion of others.  This is a subset of \emph{adaptation}, which includes all techniques that adjust a model for use on target languages, regardless of their resulting universality.} a multilingual model to a given target low-resource language, from which we take inspiration.  Among them, \citet{chau-etal-2020-parsing} augment the model's vocabulary to more effectively tokenize text, then pretrain on a small amount of data in the target language; they report significant performance improvements on a small set of low-resource languages.  In a similar vein, \citet{muller2020unseen} propose to transliterate text in the target language to Latin script to be better tokenized by the existing model, followed by additional pretraining; they observe mixed results and note that transliteration quality may be a confounding factor.  We hypothesize that these two methods can serve as the basis for improvements in modeling a broad set of low-resource languages.

In this work, we study the effectiveness, extensibility, and interaction of these two approaches to specialization: the vocabulary augmentation technique of \citet{chau-etal-2020-parsing} and the script transliteration method of \citet{muller2020unseen}.  We verify the performance of vocabulary augmentation on three tasks in a diverse set of nine low-resource languages across three different scripts, especially on non-Latin scripts (\S \ref{va}) and find that these gains are associated with improved vocabulary coverage of the target language.  We further observe a negative interaction between vocabulary augmentation and transliteration in light of a broader framework for specializing multilingual models, while noting that vocabulary augmentation offers an appealing balance of performance and cost (\S \ref{translit}).  Overall, our results highlight several possible directions for future study in the low-resource setting.  Our code, data, and hyperparameters are publicly available.\footnote{\url{https://github.com/ethch18/specializing-multilingual}}

\section{Revisiting Vocabulary Augmentation} \label{va}

We begin by revisiting the Vocabulary Augmentation method of \citet{chau-etal-2020-parsing}, which we recast more generally in light of recent work (\S \ref{va-method}).  We evaluate their claims on three different tasks, using a diverse set of languages in multiple scripts (\S \ref{va-experiments}), and find that the results hold to an even more pronounced degree in unseen low-resource languages with non-Latin scripts (\S \ref{va-results}).

\subsection{Method Overview}\label{va-method}

Following \citet{chau-etal-2020-parsing}, we consider how to apply the pretrained multilingual BERT model \citep[\mbert{};][]{devlin-etal-2019-bert} to a target low-resource language, for which both labeled and unlabeled data is scarce.  This model has produced strong CWRs for many languages \citep[\emph{inter alia}]{kondratyuk-straka-2019-75} and has been the starting model for many studies on low-resource languages \citep{muller2020unseen,pfeiffer-etal-2020-mad,wang2020extending}.  \mbert{} covers the languages with the 104 largest Wikipedias, and it uses this data to construct a wordpiece vocabulary \citep{wu2016googles} and train its transformer-based architecture \citep{Vaswani2017AttentionIA}.  Although low-resource languages are slightly oversampled, high-resource languages still dominate both the final pretraining data and the vocabulary \citep{acs-2019-exploring,devlin-etal-2019-bert}.

\citet{chau-etal-2020-parsing} note that target low-resource languages fall into three categories with respect to \mbert{}'s pretraining data: the lowest-resource languages in the data (Type 1), completely unseen low-resource languages (Type 2), and low-resouce languages with more representation (Type 0).\footnote{\citet{muller2020unseen} further subdivide Type 2 into Easy, Medium, and Hard languages, based on the performance of \mbert{} after being exposed to these languages.  However, this categorization cannot be determined \emph{a priori} for a given language.}  Due to their poor representation in the vocabulary, Type 1 and Type 2 languages achieve suboptimal tokenization and higher rates of the ``unknown'' wordpiece\footnote{The ``unknown'' wordpiece is inserted when the wordpiece algorithm is unable to segment a word-level token with the current vocabulary.} when using \mbert{} out of the box.  This hinders the model's ability to capture meaningful patterns in the data, resulting in reduced data efficiency and degraded performance.

We note that this challenge is exacerbated when modeling languages written in non-Latin scripts.  \mbert{}'s vocabulary is heavily Latin-centric \citep{acs-2019-exploring,muller2020unseen}, resulting in a significantly larger portion of non-Latin scripts being represented with ``unknown'' tokens \citep{pfeiffer2021unks} and further limiting the model's ability to generalize.  In effect, \mbert{}'s low initial performance on such languages can be attributed to its inability to represent the script itself.

To alleviate the problem of poor tokenization, \citet{chau-etal-2020-parsing} propose to specialize \mbert{} using Vocabulary Augmentation (\va{}).  Given unlabeled data in the target language, they train a new wordpiece vocabulary on the data, then select the 99 most common wordpieces in the new vocabulary that replace ``unknown'' tokens under the original vocabulary.  They then add these 99 wordpieces to the original vocabulary and continue pretraining \mbert{} on the unlabeled data for additional steps.  They further describe a tiered variant (\tva{}), in which a larger learning rate is used for the embeddings of these 99 new wordpieces.  \va{} yields strong gains over unadapted multilingual language models on dependency parsing in four low-resource languages with Latin scripts.  However, no evaluation has been performed on other tasks or on languages with non-Latin scripts, which raises our first research question:

\resq{1}: Do the conclusions of \citet{chau-etal-2020-parsing} hold for other tasks and for languages with non-Latin scripts?

We can view \va{} and \tva{} as an instantation of a more general framework of vocabulary augmentation, shared by other approaches to using \mbert{} in low-resource settings.  Given a new vocabulary $V$, number of wordpieces $n$, and learning rate multiplier $a$, the $n$ most common wordpieces in $V$ are added to the original vocabulary.  Additional pretraining is then performed, with the embeddings of the $n$ wordpieces taking on a learning rate $a$ times greater than the overall learning rate.  For \va{}, we set $n = 99$ and $a = 1$, while we treat $a$ as a hyperparameter for \tva{}.  The related E-MBERT method of \citet{wang2020extending} sets $n = \lvert V \rvert$ and $a = 1$.  Investigating various other instantiations of this framework is an interesting research direction, though it is out of the scope of this work.

\subsection{Experiments}\label{va-experiments}

We expand on the dependency parsing evaluations of \citet{chau-etal-2020-parsing} by additionally considering named entity recognition and part-of-speech tagging.  We follow \citet{kondratyuk-straka-2019-75} and compute the CWR for each token as a weighted sum of the activations at each \mbert{} layer.  For dependency parsing, we follow the setup of \citet{chau-etal-2020-parsing} and \citet{muller2020unseen} and use the CWRs as input to the graph-based dependency parser of \citet{dozat2016deep}.  For named entity recognition, the CWRs are used as input to a CRF layer, while part-of-speech tagging uses a linear projection atop the representations.  In all cases, the underlying CWRs are finetuned during downstream task training, and we do not add an additional encoder layer above the transformer outputs.  We train models on five different random seeds and report average scores and standard errors.

\subsubsection{Languages and Datasets}

\begin{table*}[!htbp]
    \centering
    \resizebox{\textwidth}{!}{%
    \begin{tabular}{llllrrrr}
        \hline
        \textbf{Language} & \textbf{Type} & \textbf{Script} & \textbf{Family} & \textbf{\# Sentences} & \textbf{\# Tokens} & \textbf{Downsample \%} & \textbf{\# WP/Token} \\
        \hline
        Bulgarian (\bg{}) & 0 & Cyrillic & Slavic & 357k & 5.6M & 10\% & 1.81 \\
        Belarusian (\be{}) & 0 & Cyrillic & Slavic & 187k & 2.7M & 10\% & 2.25 \\
        Meadow Mari (\mhr{}) & 2 & Cyrillic & Uralic & 52k & 512k & -- & 2.37 \\
        Vietnamese (\vi{}) & 0 & Latin & Viet-Muong & 338k & 6.9M & 5\% & 1.17 \\
        Irish (\ga{}) & 1 & Latin & Celtic & 274k & 5.8M & -- & 1.83 \\
        Maltese (\mt{}) & 2 & Latin & Semitic & 75k & 1.4M & -- & 2.39 \\
        Wolof (\wo{}) & 2 & Latin & Niger-Congo & 15k & 396k & -- & 1.78 \\
        Urdu (\ur{}) & 0 & Perso-Arabic & Indic & 201k & 3.6M & 20\% & 1.58 \\
        Uyghur (\ug{}) & 2 & Perso-Arabic & Turkic & 136k & 2.3M & -- & 2.54 \\
        \hline
    \end{tabular}%
    }
    \caption{Language overview and unlabeled dataset statistics: number of sentences, number of tokens, and average wordpieces per token under the original \mbert{} vocabulary.}
    \label{tab:lang_data}
\end{table*}

We select a set of nine typologically diverse low-resource languages for evaluation, including three of the original four used by \citet{chau-etal-2020-parsing}.  These languages use three different scripts and are chosen based on the availability of labeled datasets and their exemplification of the three language types identified by \citet{chau-etal-2020-parsing}.  Of the languages seen by \mbert{}, all selected Type 0 languages are within the 45 largest Wikipedias, while the remaining Type 1 languages are within the top 100.  The Type 2 languages, which are excluded from \mbert{}, are all outside of the top 150.\footnote{Based on \url{https://meta.wikimedia.org/wiki/List_of_Wikipedias}.}  Additional information about the evaluation languages is given in Tab.\ \ref{tab:lang_data}.

% https://web.archive.org/web/20201224000547/https://meta.wikimedia.org/wiki/List_of_Wikipedias

\paragraph{Unlabeled Datasets}

Following \citet{chau-etal-2020-parsing}, we use articles from Wikipedia as unlabeled data for additional pretraining in order to reflect the original pretraining data.  We downsample full articles from the largest Wikipedias to be on the order of millions of tokens in order to simulate a low-resource unlabeled setting, and we remove sentences that appear in the labeled validation or test sets.

\paragraph{Labeled Datasets}

For dependency parsing and part-of-speech tagging, we use datasets and train/test splits from Universal Dependencies \citep{nivre-etal-2020-universal}, version 2.5 \citep{zeman-etal-2019-universal}.  POS tagging uses language-specific part-of-speech tags (XPOS) to evaluate understanding of language-specific syntactic phenomena.  The Belarusian treebank lacks XPOS tags for certain examples, so we use universal part-of-speech tags instead.  Dependency parsers are trained with gold word segmentation and no part-of-speech features.  Experiments with named entity recognition use the WikiAnn dataset \citep{pan-etal-2017-cross}, following past work \citep{muller2020unseen,pfeiffer-etal-2020-mad,wu-dredze-2020-languages}.  Specifically, we use the balanced train/test splits of \citep{rahimi-etal-2019-massively}.  We note that UD datasets were unavailable for Meadow Mari, and partitioned WikiAnn datasets were missing for Wolof.

\subsubsection{Baselines}

To measure the effectiveness of \va{}, we benchmark it against unadapted \mbert{}, as well as directly pretraining \mbert{} on the unlabeled data without modifying the vocabulary \citep{chau-etal-2020-parsing,muller2020unseen,pfeiffer-etal-2020-mad}.  Following \citet{chau-etal-2020-parsing}, we refer to the latter approach as \emph{language-adaptive pretraining} (\lapt{}).  We also evaluate two monolingual baselines that are trained on our unlabeled data: fastText embeddings \citep[\fttrained{};][]{bojanowski-etal-2017-enriching}, which represent a static word vector approach; and a BERT model trained from scratch (\berttrained{}).  For \berttrained{}, we follow \citet{muller2020unseen} and train a six-layer RoBERTa model \citep{liu2019roberta} with a language-specific SentencePiece tokenizer \citep{kudo-richardson-2018-sentencepiece}.  For a fair comparison to \va{}, we use the same task-specific architectures and modify only the input representations.

\subsubsection{Implementation Details}

To pretrain \lapt{} and \va{} models, we use the code of \citet{chau-etal-2020-parsing}, who modify the pretraining code of \citet{devlin-etal-2019-bert} to only use the masked language modeling (MLM) loss.  To generate \va{} vocabularies, we train a new vocabulary of size 5000 and select the 99 wordpieces that replace the most unknown tokens.  We train with a fixed linear warmup of 1000 steps.  To pretrain \berttrained{} models, we use the HuggingFace Transformers library \citep{wolf-etal-2020-transformers}.  Following \citet{muller2020unseen}, we train a half-sized RoBERTa model with six layers and 12 attention heads.  We use a byte-pair vocabulary of size 52000 and a linear warmup of 1 epoch.  For \lapt{}, \va{}, and \berttrained{}, we train for up to 20 epochs total, selecting the highest-performing epoch based on validation masked language modeling loss.  \fttrained{} models are trained with the skipgram model for five epochs, with the default hyperparameters of \citet{bojanowski-etal-2017-enriching}.

Training of downstream parsers and taggers follows \citet{chau-etal-2020-parsing} and \citet{kondratyuk-straka-2019-75}, with an inverse square-root learning rate decay and linear warmup, and layer-wise gradual unfreezing and discriminative finetuning.  Models are trained with AllenNLP, version 2.1.0 \citep{gardner-etal-2018-allennlp}, for up to 200 epochs with early stopping based on validation performance.  We choose batch sizes to be the maximum that allows for successful training on one GPU.

\begin{table*}[!htbp]
    \begin{subtable}[h]{\textwidth}
        \centering
        \resizebox{\textwidth}{!}{%
        \begin{tabular}{lrrrrrrrrr}
            \hline
            \textbf{Rep.} & \textbf{\be{}* (0) } & \textbf{\bg{} (0)} & \textbf{\ga{} (1)} & \textbf{\mt{} (2)} & \textbf{\ug{} (2)} & \textbf{\ur{} (0)} & \textbf{\vi{} (0)} & \textbf{\wo{} (2)} & \textbf{Avg.} \\
            \hline
            \fttrained{} & 68.84 \err{7.16} & 88.86 \err{0.37} & 86.87 \err{2.55} & 89.68 \err{2.15} & \graycell{89.45 \err{1.37}} & 90.81 \err{0.31} & 81.84 \err{1.15} & 87.48 \err{0.55} & 85.48 \\
            \berttrained{} & 91.00 \err{0.30} & 94.48 \err{0.10} & 90.36 \err{0.20} & 92.61 \err{0.10} & 90.87 \err{0.13} & 89.88 \err{0.13} & 84.73 \err{0.13} & 87.71 \err{0.31} & 90.20 \\
            \hline
            \mbert{} & 94.57 \err{0.45} & 96.98 \err{0.08} & 91.91 \err{0.25} & 94.01 \err{0.17} & 78.07 \err{0.22} & 91.77 \err{0.18} & 88.97 \err{0.10} & 93.04 \err{0.20} & 91.16 \\
            \lapt{} & \textbf{95.74 \err{0.44}} & \graycell{97.15 \err{0.04}} & \graycell{93.28 \err{0.19}} & 95.76 \err{0.09} & 79.88 \err{0.27} & \graycell{92.18 \err{0.16}} & \textbf{89.64 \err{0.20}} & \textbf{94.58 \err{0.13}} & 92.28 \\
            \va{} & \graycell{95.28 \err{0.51}} & \textbf{97.20 \err{0.06}} & \textbf{93.33 \err{0.16}} & \textbf{96.33 \err{0.09}} & \textbf{91.49 \err{0.13}} & \textbf{92.24 \err{0.16}} & \graycell{89.49 \err{0.22}} & \graycell{94.48 \err{0.20}} & \textbf{93.73} \\
            \hline
        \end{tabular}%
        }
        \caption{POS tagging (accuracy).  *Belarusian uses universal POS tags.
        }
        \label{tab:pos}
    \end{subtable}
    
    \begin{subtable}[h]{\textwidth}
        \centering
        \resizebox{\textwidth}{!}{%
        \begin{tabular}{lrrrrrrrrr}
            \hline
            \textbf{Rep.} & \textbf{\be{} (0) } & \textbf{\bg{} (0)} & \textbf{\ga{} (1)} & \textbf{\mt{} (2)} & \textbf{\ug{} (2)} & \textbf{\ur{} (0)} & \textbf{\vi{} (0)} & \textbf{\wo{} (2)} & \textbf{Avg.} \\
            \hline
            \fttrained{} & 35.81 \err{2.24} & 84.03 \err{0.41} & 65.58 \err{1.21} & 68.45 \err{1.40} & 54.52 \err{1.02} & 79.33 \err{0.25} & 54.91 \err{0.79} & 70.39 \err{1.39} & 64.13 \\
            \berttrained{} & 45.77 \err{1.35} & 84.61 \err{0.27} & 64.02 \err{0.49} & 65.92 \err{0.45} & 60.34 \err{0.27} & 78.07 \err{0.22} & 54.70 \err{0.27} & 60.12 \err{0.39} & 64.19 \\
            \hline
            \mbert{} & 71.83 \err{0.90} & \graycell{91.62 \err{0.23}} & 71.68 \err{0.62} & 76.63 \err{0.35} & 47.70 \err{0.44} & \graycell{81.45 \err{0.26}} & 64.58 \err{0.42} & 76.24 \err{0.83} & 72.72 \\
            \lapt{} & \graycell{72.77 \err{1.12}} & \textbf{92.08 \err{0.31}} & \textbf{74.79 \err{0.12}} & 81.53 \err{0.37} & 50.67 \err{0.34} & \graycell{81.78 \err{0.44}} & \textbf{66.15 \err{0.41}} & \textbf{80.34 \err{0.14}} & 75.01 \\
            \va{} & \textbf{73.22 \err{1.23}} & \graycell{91.90 \err{0.20}} & 74.35 \err{0.22} & \textbf{82.00 \err{0.31}} & \textbf{67.55 \err{0.17}} & \textbf{81.88 \err{0.25}} & \graycell{65.64 \err{0.12}} & \graycell{80.22 \err{0.41}} & \textbf{77.09} \\
            \hline
        \end{tabular}%
        }
        \caption{UD parsing (LAS).
        }
        \label{tab:ud}
    \end{subtable}
    
    \begin{subtable}[h]{\textwidth}
        \centering
        \resizebox{\textwidth}{!}{%
        \begin{tabular}{lrrrrrrrrr}
            \hline
            \textbf{Rep.} & \textbf{\be{} (0) } & \textbf{\bg{} (0)} & \textbf{\ga{} (1)} & \textbf{\mt{} (2)} & \textbf{\ug{} (2)} & \textbf{\ur{} (0)} & \textbf{\vi{} (0)} & \textbf{\mhr{} (2)} & \textbf{Avg.} \\
            \hline
            \fttrained{} & 84.26 \err{0.86} & 87.98 \err{0.76} & 67.21 \err{4.30} & 33.53 \err{17.89} & -- & \graycell{92.85 \err{2.04}} & 85.57 \err{1.98} & 35.28 \err{13.81} & 60.84 \\
            \berttrained{} & 88.08 \err{0.62} & 90.31 \err{0.20} & 76.58 \err{0.98} & 54.64 \err{3.51} & \graycell{61.54 \err{3.70}} & 94.04 \err{0.55} & 88.08 \err{0.15} & 54.17 \err{2.88} & 75.93 \\
            \hline
            \mbert{} & \graycell{91.13 \err{0.07}} & 92.56 \err{0.09} & 82.82 \err{0.57} & 61.86 \err{2.60} & 50.76 \err{1.86} & 94.60 \err{0.34} & \graycell{92.13 \err{0.27}} & \graycell{61.85 \err{3.25}} & 78.46 \\
            \lapt{} & \textbf{91.61 \err{0.74}} & \textbf{92.96 \err{0.13}} & \graycell{84.13 \err{0.78}} & \textbf{81.53 \err{2.33}} & 56.76 \err{4.91} & 95.17 \err{0.29} & \graycell{92.41 \err{0.15}} & \graycell{59.17 \err{5.15}} & 81.72 \\
            \va{} & \graycell{91.38 \err{0.56}} & 92.70 \err{0.11} & \textbf{84.82 \err{1.00}} & \graycell{80.00 \err{2.77}} & \textbf{68.93 \err{3.30}} & \textbf{95.43 \err{0.22}} & \textbf{92.43 \err{0.16}} & \textbf{64.23 \err{3.07}} & \textbf{83.74} \\
            \hline
        \end{tabular}%
        }
        \caption{NER (macro F1).  -- indicates that a model did not converge.
        }
        \label{tab:ner}
    \end{subtable}
    
    \caption{Results on POS tagging, UD parsing, and NER, with standard deviations from five random initializations.  \textbf{Bolded} results are the maximum for each language, and scores in \graybox{gray} are not significantly worse than the best model (1-sided paired $t$-test, $p = 0.05$ with Bonferonni correction).}
    \label{tab:all}
\end{table*}

\subsection{Results}\label{va-results}

Tab.\ \ref{tab:all} presents performance of the different input representations on POS tagging, dependency parsing, and named entity recognition.  \va{} achieves strong results across all languages and tasks and is the top performer in the majority of them, suggesting that augmenting the vocabulary addresses \mbert{}'s limited vocabulary coverage of the target language and is beneficial during continued pretraining.

The relative gains that \va{} provides appear to correlate not only with language type, as in the findings of \citet{chau-etal-2020-parsing}, but also with each language's script.  For instance, in Vietnamese, which is a Type 0 Latin script language, the improvements from \va{} are marginal at best, reflecting the Latin-dominated pretraining data of \mbert{}.  Irish, the Type 1 Latin script language, is only slightly more receptive.  However, Type 0 languages in Cyrillic and Arabic scripts, which are less represented in \mbert{}'s pretraining data, are more receptive to \va{}, with \va{} even outperforming all other methods for Urdu.  This trend is amplified in the Type 2 languages, as the improvements for Maltese and Wolof are small but significant.  However, they are dwarfed in magnitude by those of Uyghur, where \va{} achieves up to a 57\% relative error reduction over \lapt{}.

This result corroborates the findings of both \citet{chau-etal-2020-parsing} and \citet{muller2020unseen} and answers \resq{1}.  Prior to specialization, \mbert{} is especially poorly equipped to handle unseen low-resource languages and languages in non-Latin scripts due to its inability to model the script itself.  In such cases, specialization via \va{} is beneficial, providing \mbert{} with explicit signal about the target language and script while maintaining its language-agnostic insights.  On the other hand, this also motivates additional investigation into remedies for the script imbalance at a larger scale, e.g., more diverse pretraining data.

\subsection{Analysis}

We perform further analysis to investigate \va{}'s patterns of success.  Concretely, we hypothesize that \va{} significantly improves the tokenizer's coverage of target languages where it is most successful.  Inspired by \citet{acs-2019-exploring}, \citet{chau-etal-2020-parsing}, and \citet{rust-etal-2021-good}, we quantify tokenizer coverage using the percentage of tokens in the raw text that yield unknown wordpieces when tokenized with a given vocabulary (``UNK token percentage'').  These are tokens whose representations contain at least partial ambiguity due to the inclusion of the unknown wordpiece.

Tab.~\ref{tab:main-deltas} presents the UNK token percentage for each dataset using the \mbert{} vocabulary, averaged over each script and language type.  This vocabulary is used in \lapt{} and represents the baseline level of vocabulary coverage.  We also include the change in the UNK token percentage between the \mbert{} and \va{} vocabularies, which quantifies the coverage improvement.  Both sets of values are juxtaposed against the average change in task-specific performance from \lapt{} to \va{}, representing the effect of augmenting the vocabulary on task-specific performance.

We observe that off-the-shelf \mbert{} already attains relatively high vocabulary coverage for Type 0 and 1 languages, as well as languages written in Latin and Cyrillic scripts.  On the other hand, up to one-fifth of the tokens in Arabic languages and one-sixth of those in Type 2 languages yield an unknown wordpiece.  For these languages, there is great room for increasing tokenizer coverage, and \va{} indeed addresses this more tangible need.  This aligns with the task-specific performance improvements for each group and helps to explain our results in \S \ref{va-results}.

\begin{table*}[!htbp]
    \centering
    \resizebox{\textwidth}{!}{%
    \begin{tabular}{lrrrrrrrrr}
        \hline
        \textbf{Lang. Group} & \multicolumn{3}{c}{\textbf{Avg. UNK Token \% (\mbert{})}} & \multicolumn{3}{c}{\textbf{Avg. UNK Token \% ($\Delta$)}} & \multicolumn{3}{c}{\textbf{Avg. Task Performance ($\Delta$)}} \\
        \textbf{(\# of Langs.)} & Unlabeled & UD & WikiAnn & Unlabeled & UD & WikiAnn & POS & UD & NER \\
        \hline
        All (9) & 5.9 \% \neutd{} & 5.2 \% \neutd{} & 6.2 \% \neutd{} & --5.3 \% \neutd{} & 4.7 \% \neutd{} & --5.8 \% \neutd{} & +1.45 \neutd{} & +2.08 \neutd{} & +2.02 \neutd{} \\
        \hline
        Type 0 (4) & 1.0 \% \decdd{} & 0.3 \% \decdd{} & 1.2 \% \decdd{} & --0.9 \% \incdd{} & --0.3 \% \incdd{} & --1.2 \% \incdd{} & --0.13 \decd{} & --0.04 \decd{} & --0.05 \decd{} \\
        Type 1 (1) & 0.3 \% \decdd{} & 0.0 \% \decdd{} & 0.4 \% \decdd{} & --0.3 \% \incdd{} & --0.00 \% \incdd{} & --0.4 \% \incdd{} & +0.05 \decd{} & --0.44 \decd{} & +0.69 \decd{} \\
        Type 2 (4) & 12.3 \% \incdd{} & 13.5 \% \incdd{} & 14.8 \% \incdd{} & --10.8 \% \decdd{} & --12.1 \% \decdd{} & --13.7 \% \decdd{} & +4.03 \incd{} & +5.74 \incd{} & +5.23 \incd{} \\
        \hline
        Latin (4) & 1.2 \% \decdd{} & 0.6 \% \decdd{} & 2.4 \% \decdd{} & --1.2 \% \incdd{} & --0.6 \% \incdd{} & --2.3 \% \incdd{} & +0.09 \decd{} & --0.15 \decd{} & --0.27 \decd{} \\
        Cyrillic (3) & 3.6 \% \decdd{} & 0.6 \% \decdd{} & 2.8 \% \decdd{} & --3.6 \% \incdd{} & --0.6 \% \incdd{} & --2.7 \% \incdd{} & --0.21 \decd{} & +0.14 \decd{} & +1.52 \decd{} \\
        Arabic (2) & 19.0 \% \incdd{} & 19.2 \% \incdd{} & 16.9 \% \incdd{} & --16.1 \% \decdd{} & --17.0 \% \decdd{} & --15.5 \% \decdd{} & +5.84 \incd{} & +8.49 \incd{} & +6.22 \incd{} \\
        \hline
    \end{tabular}%
    }
    \caption{Average UNK token percentage under the \mbert{} vocabulary (left); change in UNK token percentage from \mbert{} to \va{} vocabularies (center); and average task performance change from \lapt{} to \va{} (right).  Averages are computed overall and within each script and language type, with comparisons to the overall average; all UNK token percentages are computed on the respective training sets for illustration.  Note that Uyghur accounts for a large portion of the behavior of the Type 2/Arabic rows.
    }
    \label{tab:main-deltas}
\end{table*}
It is notable that \va{} does not always eliminate the issue of unknown wordpieces, even in languages for which \mbert{} attains high vocabulary coverage.  This suggests that the remaining unknown wordpieces in these languages are more sparsely distributed (i.e., they represent low frequency sequences), while the unknown wordpieces in languages with lower vocabulary coverage represent sequences that occur more commonly.  As a result, augmenting the vocabulary in such languages quickly improves coverage while associating these commonly occurring sequences with each other, which benefits the overall tokenization quality.

\begin{figure*}
    \centering
    \begin{subfigure}[b]{0.32\textwidth}
        \centering
        \includegraphics[trim={0.1cm 0.1cm 0.1cm 0.1cm}, clip, width=\textwidth]{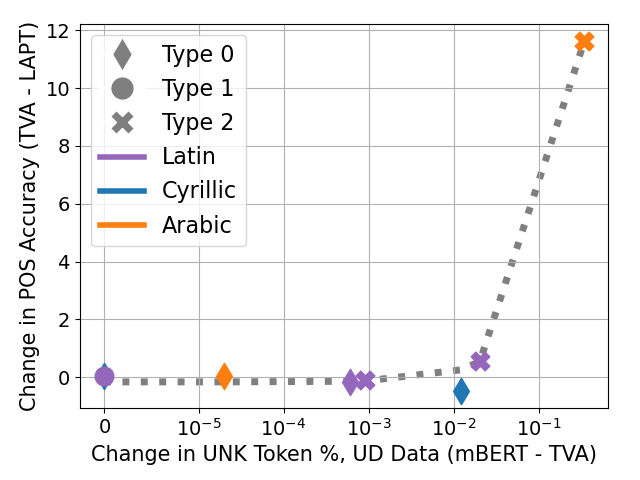}
        \caption{POS tagging.}
    \end{subfigure}
    \hfill
    \begin{subfigure}[b]{0.32\textwidth}
        \centering
        \includegraphics[trim={0.1cm 0.1cm 0.1cm 0.1cm}, clip, width=\textwidth]{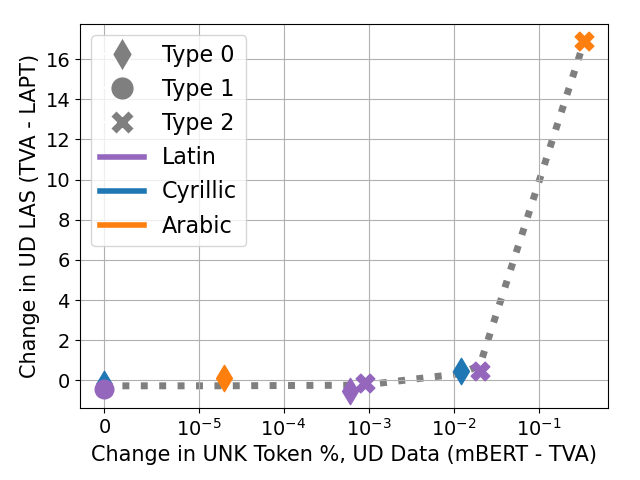}
        \caption{UD parsing.}
    \end{subfigure}
    \hfill
    \begin{subfigure}[b]{0.32\textwidth}
        \centering
        \includegraphics[trim={0.1cm 0.1cm 0.1cm 0.1cm}, clip, width=\textwidth]{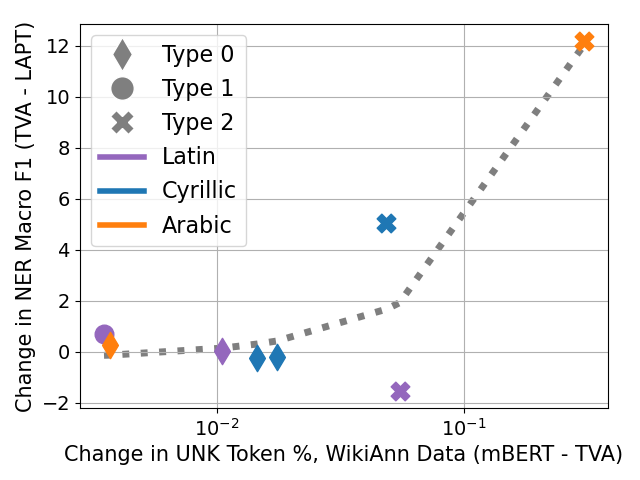}
        \caption{NER.}
    \end{subfigure}
    \caption{Relationship between the change in UNK token percentage on task data and the change in task performance, from (\mbert{}/\lapt{} to \va{}), with a 1-degree line of best fit.  All vocabulary values are computed on the respective training sets.
    }
    \label{fig:main-deltas}
\end{figure*}

We further explore the association between the improvements in vocabulary coverage and task-specific performance in Fig.~\ref{fig:main-deltas}.  Although we do not find that languages from the same types or scripts form clear clusters, we nonetheless observe a loose correlation between the two factors in question and see that \va{} delivers greater performance gains on Type 2 and Arabic-script languages compared to their Type 0/1 and Latin-script counterparts, respectively.  To quantify the strength of this association, we also compute the language-level Spearman correlation between the change in UNK token percentage on the unlabeled dataset\footnote{We benchmark against the unlabeled dataset instead of task-specific ones for comparability.} from the \mbert{} to \va{} vocabulary and the task-specific performance improvements from \lapt{} to \va{}.  The resulting $\rho$-values -- 0.29 for NER, 0.56 for POS tagging, and 0.81 for UD parsing -- suggest that this set of factors is meaningful for some tasks, though additional and more fine-grained analysis in future work should give a more complete explanation.

\section{Mix-in Specialization: \va{} and Transliteration} \label{translit}

We now expand on the observation made in \S \ref{va-results} regarding the difficulties that \mbert{} encounters when faced with unseen low-resource languages in non-Latin scripts because of its inability to model the script.  Having observed that \va{} is beneficial in such cases, we now investigate the interaction between this method and another specialization approach that targets this problem.  Specifically, we consider the transliteration  methods of \citet{muller2020unseen}, in which unseen low-resource languages in non-Latin scripts are transliterated into the Latin script, often using transliteration schemes inspired by the Latin orthographies of languages related to the target language.  They hypothesize that the increased similarity in the languages' writing systems, combined with \mbert{}'s overall Latin-centricity, provides increased opportunity for crosslingual transfer.

We can view transliteration as a inverted form of vocabulary augmentation: instead of adapting the model to the needs of the data, the data is adjusted to meet the assumptions of the model.  Furthermore, the transliteration step is performed prior to pretraining \mbert{} on additional unlabeled data in the target language, the same stage at which \va{} is performed.  In both cases, the ultimate goal is identical: improving tokenization and more effectively using available data.  We can thus view transliteration and \va{} as two instantiations of a more general \emph{mix-in} paradigm for model specialization, whereby various transformations (mix-ins) are applied to the data and/or model prior to performing additional pretraining.  These mix-ins target different components of the experimental pipeline, which naturally raises our second research question:

\resq{2}: How do the \va{} and transliteration mix-ins for \mbert{} compare and interact?

\subsection{Method and Experiments}
To test this research question, we apply transliteration and \va{} in succession and evaluate their compatibility.  Given unlabeled data in the target language, we first transliterate it into Latin script, which decreases but does not fully eliminate the issue of unseen wordpieces.  We then perform \va{}, generating the vocabulary for augmentation based on the transliterated data.

We evaluate on Meadow Mari and Uyghur, which are Type 2 languages where transliteration was successfully applied by \citet{muller2020unseen}.  To transliterate the data, we use the same methods as \citet{muller2020unseen}: Meadow Mari uses the \texttt{transliterate}\footnote{\url{https://pypi.org/project/transliterate}} package, while Uyghur uses a linguistically-motivated transliteration scheme\footnote{\url{https://github.com/benjamin-mlr/mbert-unseen-languages}} aimed at associating Uyghur with Turkish.  We use the same training scheme, model architectures, and baselines as in \S \ref{va-experiments}, the only difference being the use of transliterated data.  This includes directly pretraining on the unlabeled data (\lapt{}), which is comparable to the highest-performing transliteration models of \citet{muller2020unseen}.  Although our initial investigation of \va{} in \S \ref{va} also included non-Type 2 languages of other scripts, we omit them from our investigation based on the finding of \citet{muller2020unseen} that transliterating higher-resource languages into Latin scripts is not beneficial.

\subsection{Results}

Tab.\ \ref{tab:all-translit} gives the results of our transliteration mix-in experiments.  For the \mbert{}-based models, both \va{} and transliteration provide strong improvements over their respective baselines.  Specifically, the improvements from \lapt{} to \va{} and \lapt{} to \lapt{} with transliteration are most pronounced.  This verifies the independent results of \citet{chau-etal-2020-parsing} and \citet{muller2020unseen} and suggests that in the non-Latin low-resource setting, unadapted additional pretraining is insufficient, but that the mix-in stage between initial and additional pretraining is amenable to performance-improving modifications.  Unsurprisingly, transliteration provides no consistent improvement to the monolingual baselines, since the noisy transliteration process removes information without improving crosslingual alignment.

However, \va{} and transliteration appear to interact negatively.  Although \va{} with transliteration improves over plain \va{} for Uyghur POS tagging and dependency parsing, it still slightly underperforms \lapt{} with transliteration for the latter.  For the two NER experiments, \va{} with transliteration lags both methods independently.  One possible explanation is that transliteration into Latin script serves as implicit vocabulary augmentation, with embeddings that have already been updated during the initial pretraining stage; as a result, the two sources of augmentation conflict.  Alternatively, since the transliteration process merges certain characters that are distinct in the original script, \va{} may augment the vocabulary with misleading character clusters.  Either way, additional vocabulary augmentation is generally not as useful when combined with transliteration, answering \resq{2}.

Nonetheless, additional investigation into the optimal amount of vocabulary augmentation might yield a configuration that is consistently complementary to transliteration and is an interesting direction for future work.  Furthermore, designing linguistically-informed transliteration schemes like those devised by \citet{muller2020unseen} for Uyghur requires large amounts of time and domain knowledge.  \va{}'s fully data-driven nature and relatively comparable performance suggest that it achieves an appealing balance between performance gain and implementation difficulty.

\begin{table*}[!htbp]
    \centering
    \resizebox{\textwidth}{!}{%
    \begin{tabular}{lrrrrrrrr}
        \hline
        \textbf{Rep.} & \multicolumn{2}{c}{\textbf{\mhr{} (NER)}} & \multicolumn{2}{c}{\textbf{\ug{} (NER)}} & \multicolumn{2}{c}{\textbf{\ug{} (POS)}} & \multicolumn{2}{c}{\textbf{\ug{} (UD)}} \\
        \hline
        \fttrained{} & 35.28 $\to$ 41.32 & \del{+6.04} & \multicolumn{2}{c}{--} & 89.45 $\to$ 89.03 & \del{--0.42} & 54.52 $\to$ 54.45 & \del{--0.07} \\
        \berttrained{} & 54.17 $\to$ 48.45 & \del{--5.72} & 61.54 $\to$ 63.05 & \del{+1.51} & 90.87 $\to$ 90.76 & \del{--0.09} & 60.34 $\to$ 60.08 & \del{--0.26}\\
        \hline
        \mbert{} & 61.85 $\to$ 63.84 & \del{+1.99} & 50.76 $\to$ 56.80 & \del{+6.04} & 78.07 $\to$ 91.34 & \del{+13.27}  & 47.70 $\to$ 65.85 & \del{+18.15}\\
        \lapt{} & 59.17 $\to$ 63.68 & \del{+4.51} & 56.76 $\to$ 67.57 & \del{+10.81} & 79.88 $\to$ 92.59 & \del{+12.71} & 50.67 $\to$ \textbf{69.39} & \del{+18.72}\\
        \va{} & \textbf{64.23} $\to$ 63.19 & \del{--1.04} & \textbf{68.93} $\to$ 67.10 & \del{--1.83} & 91.49 $\to$ \textbf{92.64} & \del{+1.15} & 67.55 $\to$ 68.58 & \del{+1.03} \\
        \hline
    \end{tabular}%
    }
    \caption{Comparison of model performance before and after transliteration.  \textbf{Bolded} results are the maximum for each language-task pair.  -- indicates that a model did not converge.}
    \label{tab:all-translit}
\end{table*}

\section{Related Work}

Our work follows a long line of studies investigating the performance of multilingual language models like \mbert{} in various settings.  The exact source of such models' crosslingual ability is contested: early studies attributed \mbert{}'s success to vocabulary overlap between languages \citep{Cao2020Multilingual,pires-etal-2019-multilingual,wu-dredze-2019-beto}, but subsequent studies find typological similarity and parameter sharing to be better explanations \citep{conneau-etal-2020-emerging,wang2019cross}.  Nonetheless, past work has consistently highlighted the limitations of multilingual models in the context of low-resource languages.  \citet{conneau-etal-2020-unsupervised} highlight the tension between crosslingual transfer and per-language model capacity, which poses a challenge for low-resource languages that require both.  Indeed, \citet{wu-dredze-2020-languages} find that \mbert{} is unable to outperform baselines in the lowest-resource seen languages.  Our experiments build off these insights, which motivate the development of methods for adapting \mbert{} to target low-resource languages.  

\paragraph{Adapting Language Models}
Several prior studies have proposed methods for adapting pretrained models to a downstream task.  The simplest of these is to perform additional pretraining on unlabeled data in the target language \citep{chau-etal-2020-parsing,muller2020unseen,pfeiffer-etal-2020-mad}, which in turn builds off similar approaches for domain adaptation \citep{gururangan-etal-2020-dont,han-eisenstein-2019-unsupervised}.  Recent work uses one or more of these additional pretraining stages to specifically train modular adapter layers for specific tasks or languages, with the goal of maintaining a language-agnostic model while improving performance on individual languages \citep{pfeiffer-etal-2020-mad,pfeiffer-etal-2021-adapterfusion,vidoni2020orthogonal}.  However, as \citet{muller2020unseen} note, the typological diversity of the world's languages ultimately limits the viability of this approach.

On the other hand, many adaptation techniques have focused on improving representation of the target language by modifying the model's vocabulary or tokenization schemes \citep{chung-etal-2020-improving,clark2021canine,wang-etal-2021-multi-view}.  This is well-motivated: \citet{artetxe-etal-2020-cross} emphasize representation in the vocabulary as a key factor for effective crosslingual transfer, while \citet{rust-etal-2021-good} find that \mbert{}'s tokenization scheme for many languages is subpar.  \citet{pfeiffer2021unks} further observe that for languages with unseen scripts, a large proportion of the language is mapped to the generic ``unknown'' wordpiece, and they propose a matrix factorization-based approach to improve script representation.  \citet{wang2020extending} extend \mbert{}'s vocabulary with an entire new vocabulary in the target language to facilitate zero-shot transfer to low-resource languages from English.  The present study most closely derives from \citet{chau-etal-2020-parsing}, who select 99 wordpieces with the greatest amount of coverage to augment \mbert{}'s vocabulary while preserving the remainder; and \citet{muller2020unseen}, who transliterate target language data into Latin script to improve vocabulary coverage.  We deliver new insights on the effectiveness and applicability of these methods.

\section{Conclusion}
We explore the interactions between vocabulary augmentation and script transliteration for specializing multilingual contextual word representations in low-resource settings.  We confirm vocabulary augmentation's effectiveness on multiple languages, scripts, and tasks; identify the mix-in stage as amenable to specialization; and observe a negative interaction between vocabulary augmentation and script transliteration.  Our findings highlight several open questions in model specialization and low-resource natural language processing at large, motivating further study in this area.

Future directions for investigation are manifold.  In particular, our results in this work unify the separate findings of past works, which use \mbert{} as a case study; a natural continuation would extend these methods to a broader set of multilingual models, such as mT5 \citep{xue-etal-2021-mt5} and XLM-R \citep{conneau-etal-2020-unsupervised}, in order to obtain a clearer understanding of the factors behind specialization methods' patterns of success.  While we intentionally choose a set of small unlabeled datasets to evaluate on a setting applicable to the vast majority of the world's low-resource languages, we acknowledge great variation in the amount of  unlabeled data available in different languages.  Continued study on the applicability of these methods to datasets of different sizes is an important future step.  An interesting direction of work is to train multilingual models on data where script respresentation is more balanced, which might also allow for different output scripts for transliteration.  Given that the mix-in stage is an effective opportunity to specialize models to target languages, constructing mix-ins at both the data and model level that are complementary by design has potential to be beneficial. Finally, future work might shed light on the interaction between different configurations of the adaptations studied here (e.g., the number of wordpiece types used in vocabulary augmentation).

\section*{Acknowledgments}

We thank Jungo Kasai, Phoebe Mulcaire, and members of UW NLP for their helpful comments on preliminary versions of this paper.  We also thank Benjamin Muller for insightful discussions and providing details about transliteration methods and baselines.  Finally, we thank the anonymous reviewers for their helpful remarks.

\bibliography{references}
\bibliographystyle{acl_natbib}

\end{document}